\documentclass[10pt,twocolumn,letterpaper]{article}

\usepackage{cvpr}              

\usepackage{bm}
\usepackage{multirow}
\usepackage{adjustbox}
\usepackage{makecell}

\usepackage{array}

\newcommand*\rot{\multicolumn{1}{R{45}{2.5em}}}
\newcolumntype{R}[2]{>{\adjustbox{angle=#1,lap=\width-(#2)}\bgroup}l<{\egroup}}
\usepackage{multirow}
\usepackage{dsfont}
\usepackage{wasysym}

\definecolor{cvprblue}{rgb}{0.21,0.49,0.74}
\usepackage[pagebackref,breaklinks,colorlinks,allcolors=cvprblue]{hyperref}

\title{BEVDiffuser: Plug-and-Play Diffusion Model for BEV Denoising \\ with Ground-Truth Guidance}

\author{
Xin Ye, \,\,  
Burhaneddin Yaman\thanks{Corresponding author.}, \,\,  
Sheng Cheng, \,\,  
Feng Tao, \,\,   
Abhirup Mallik, \,\,  
Liu Ren \\ 
Bosch Research North America \& Bosch Center for Artificial Intelligence (BCAI) \ \ \ \\ 
 {\tt\small \{xin.ye3, burhaneddin.yaman, sheng.cheng, feng.tao2, abhirup.mallik, liu.ren\}@us.bosch.com} \\
 {\small\url{https://xin-ye-1.github.io/BEVDiffuser}}
}
\begin{document}
\maketitle
\begin{abstract}
Bird's-eye-view (BEV) representations play a crucial role in autonomous driving tasks. Despite recent advancements in BEV generation, inherent noise, stemming from sensor limitations and the learning process, remains largely unaddressed, resulting in suboptimal BEV representations that adversely impact the performance of downstream tasks. To address this, we propose BEVDiffuser, a novel diffusion model that effectively denoises BEV feature maps using the ground-truth object layout as guidance.  BEVDiffuser can be operated in a plug-and-play manner during training time to enhance existing BEV models without requiring any architectural modifications. Extensive experiments on the challenging nuScenes dataset demonstrate BEVDiffuser's exceptional denoising and generation capabilities, which enable significant enhancement to existing BEV models, as evidenced by notable improvements of 12.3\% in mAP and 10.1\% in NDS achieved for 3D object detection without introducing additional computational complexity. Moreover, substantial improvements in long-tail object detection and under challenging weather and lighting conditions further validate BEVDiffuser's effectiveness in denoising and enhancing BEV representations.
\end{abstract}    
\section{Introduction}
\label{sec:intro}

\begin{figure}[ht!]
    \small
    \centering
     \begin{subfigure}[b]{0.15\textwidth}
         \centering
         \includegraphics[width=\textwidth]{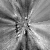}
         \caption{BEVFormer \cite{bevformer}}
         \label{fig:bev_tiny}
     \end{subfigure}
     \begin{subfigure}[b]{0.15\textwidth}
         \centering
         \includegraphics[width=\textwidth]{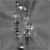}
         \caption{BEVDiffuser}
         \label{fig:bev_ours}
     \end{subfigure} 
     \begin{subfigure}[b]{0.15\textwidth}
         \centering
         \includegraphics[width=\textwidth]{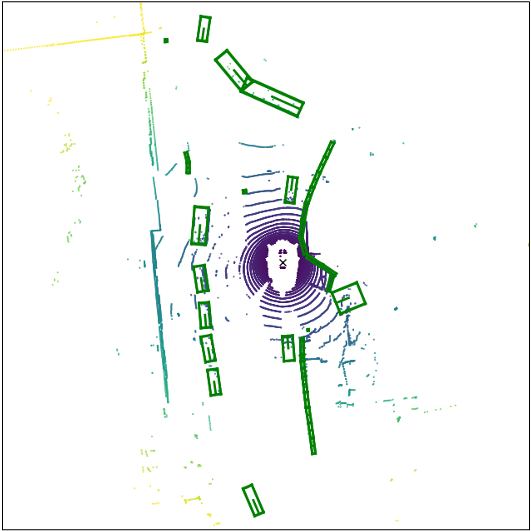}
         \caption{LiDAR Top View }
         \label{fig:gt_view}
     \end{subfigure}
    \caption{Comparisons of BEV feature maps: (a) generated by BEVFormer (tiny) \cite{bevformer}, (b) denoised by BEVDiffuser in $5$ steps. Channel-wise features are averaged for visualization. BEVDiffuser denoises and substantially enhances the BEV feature maps.}
    \vspace{-5pt}
    \label{fig:bev}
\end{figure}

Bird's-eye-view (BEV) representations have become crucial in advancing autonomous driving tasks, including perception, prediction, and planning, by providing a comprehensive top-down understanding of the surrounding environment \cite{bevformer,liu2023bevfusion,Uniad,VAD}. By integrating data from various sensors, such as multi-view cameras and LiDAR, BEV generates a unified scene representation that empowers autonomous systems to make more accurate and informed decisions. The effectiveness of the BEV representations has sparked considerable interest, resulting in a diverse set of approaches for learning BEV representations from single-modal \cite{yin2021center, bevformer} or multi-modal \cite{liu2023bevfusion, lin2024rcbevdet} sensors, using geometry-based \cite{philion2020lift} or transformer-based \cite{bevformer} methods. These advanced BEV generation techniques have emerged as state-of-the-art solutions for a variety of benchmark tasks, including 3D object detection \cite{liu2023bevfusion, huang2022bevdet4d}, map segmentation \cite{philion2020lift, pan2020cross} and autonomous planning \cite{Uniad,VAD}.

Despite recent advancements in BEV generation, the issue of noise in these BEV representations remains largely unresolved. Generated BEV representations are inherently noisy (see Fig.~\ref{fig:bev_tiny}) due to the imperfections of acquisition sensors such as camera and LiDAR, as well as the limitations in the learning process \cite{zou2024diffbev,le2024diffuser}. The noise from acquisition sensors introduces inaccuracies, including imprecise localization of object boundaries in BEV feature maps, which degrades performance in downstream tasks. Additionally, in the absence of direct supervision, BEV representations are typically optimized only for downstream task performance, leading to potential biases within the BEV feature maps. Generative models, particularly diffusion models, are well-suited to address this challenge due to their powerful denoising capabilities \cite{sohl2015deep, songdenoising, rombach2022high}. Diffusion models have demonstrated remarkable success in image and video generation \cite{ramesh2022hierarchical,rombach2022high,blattmann2023align}, and recent studies have extended their applicability to tasks such as image classification and object detection \cite{li2023your,chen2023diffusiondet, nachkov2023diffusion}. Leveraging diffusion models to denoise and enhance BEV representations holds significant potential for improving the robustness and accuracy of BEV-based downstream tasks.

In this work, we introduce BEVDiffuser, a novel diffusion model that denoises BEV representations with ground-truth guidance. BEVDiffuser is trained on BEV feature maps generated by existing BEV models, such as BEVFormer and BEVFusion~\cite{bevformer,liu2023bevfusion}. We add varying levels of noise to these BEV feature maps and train BEVDiffuser to predict the clean BEV, conditioned on the ground-truth object layout to effectively guide the denoising process. Once trained, BEVDiffuser operates in a plug-and-play manner, enhancing current BEV models by providing denoised BEV feature maps as additional supervision during training. BEVDiffuser is used only in training time and removed at deployment, leaving the enhanced BEV models without any architectural modifications for inference. Consequently, BEVDiffuser improves the performance of existing BEV models without requiring any adaptation efforts or introducing any computational latency at inference time.

BEVDiffuser, as a flexible plug-and-play module, can be seamlessly incorporated into any BEV model. In this study, we conduct an extensive evaluation of BEVDiffuser on four widely adopted state-of-the-art BEV models using the challenging nuScenes~\cite{nuscenes} dataset. The experimental results demonstrate BEVDiffuser's exceptional denoising capabilities (see Fig.~\ref{fig:bev_ours}), which enable significant enhancements to existing BEV models, demonstrated by notable improvements of $12.3\%$ in mAP and $10.1\%$ in NDS for 3D object detection. Additionally, our experiments show that BEVDiffuser substantially improves performance in long-tail object detection and under challenging weather and lighting conditions, highlighting its ability to produce more accurate and robust BEV representations. Furthermore, BEVDiffuser also shows high-quality BEV generation capabilities from pure noise with layout conditioning, which can pave the way for large-scale data collection to advance autonomous driving. Qualitative visualizations further validate the observed quantitative improvements. 

We summarize our main contributions as follows:
\begin{itemize}
    \item We propose BEVDiffuser, a novel diffusion model that effectively denoises BEV feature maps using the ground-truth object layout as guidance. 
    \item BEVDiffuser can be operated in a plug-and-play manner during training time to enhance existing BEV models without modifying their architectures or introducing additional computational overhead during inference.
    \item Extensive experiments on the nuScenes dataset demonstrate that BEVDiffuser possesses strong BEV denoising and generation capabilities, significantly enhances BEV models both quantitatively and qualitatively, and exhibits improved robustness in long-tail cases and adverse weather and lighting conditions.   
\end{itemize}

\section{Related Work}
\label{sec:related_work}
\subsection{BEV Feature Map}
\label{sec:bev}
Camera-only BEV feature generation works can be broadly categorized into two main approaches: geometry-based methods, represented by Lift-Splat-Shoot (LSS) \cite{philion2020lift}, and transformer-based methods, exemplified by BEVFormer \cite{bevformer}. LSS \cite{philion2020lift} generates BEV feature maps from multi-view images by leveraging the estimated depth distribution, followed by \cite{li2023bevdepth, huang2021bevdet, huang2022bevdet4d}. In contrast, transformer-based methods utilize powerful attention mechanism to extract attended image features for BEV generation. BEVFormer \cite{bevformer} and its follow-up BEVFormerV2 \cite{bevformerv2} have gained significant interest as they capture both spatial and temporal information through spatial cross-attention and temporal self-attention mechanisms, respectively. Another line of work presents strategies to fuse multi-modal sensor inputs for more robust BEV feature generation \cite{liang2022bevfusion,liu2023bevfusion,lin2024rcbevdet}. BEVFusion \cite{liu2023bevfusion} is a representative work that introduces a unified framework for camera and LiDAR sensors by combining multi-modal features in BEV space. In contrast to these works, we propose a plug-and-play diffusion model designed to enhance the BEV feature maps by denoising the intrinsic noise from both the acquisition sensors and the learning process.

\subsection{Diffusion Model Enhanced BEV} 
Diffusion models are a class of generative models that have demonstrated impressive performance and stability \cite{sohl2015deep,song2019generative, ho2020denoising}. While diffusion models have been primarily used for generative tasks, such as image generation \cite{rombach2022high, zhang2023adding, ramesh2022hierarchical,zheng2023layoutdiffusion} and video generation \cite{singer2023makeavideo, blattmann2023align, wu2024lamp, wang2024videocomposer}, their applications to downstream tasks such as image classification \cite{li2023your}, object detection \cite{chen2023diffusiondet}, semantic segmentation \cite{li2023open}, and motion prediction \cite{jiang2023motiondiffuser} have recently been investigated. 

Only a few approaches have been proposed to use diffusion models for enhancing the BEV feature maps \cite{le2024diffuser, zou2024diffbev}, which is the focus of this study. Specifically, DiffBEV \cite{zou2024diffbev} applies a conditional diffusion model to progressively refine the noisy BEV feature maps, using the learned features as conditions. The denoised BEV is then fused with the original BEV to perform downstream tasks. Similarly, DifFUSER \cite{le2024diffuser} leverages a diffusion model for better sensor fusion. It enhances the fused features obtained from camera and LiDAR sensors by denoising them conditioned on partial camera and LiDAR features during run time. Both approaches demonstrate the potential of diffusion models for denoising and enhancing BEV feature maps.  However, unlike our BEVDiffuser, these approaches rely on noisy information as conditions to guide the denoising process which is less effective. Moreover, they require multiple passes through their integrated diffusion model during inference, making them computationally expensive for latency-critical real-world applications like autonomous driving.
 
\begin{figure*}[ht!]
    \centering
    \includegraphics[width=1.0\textwidth]{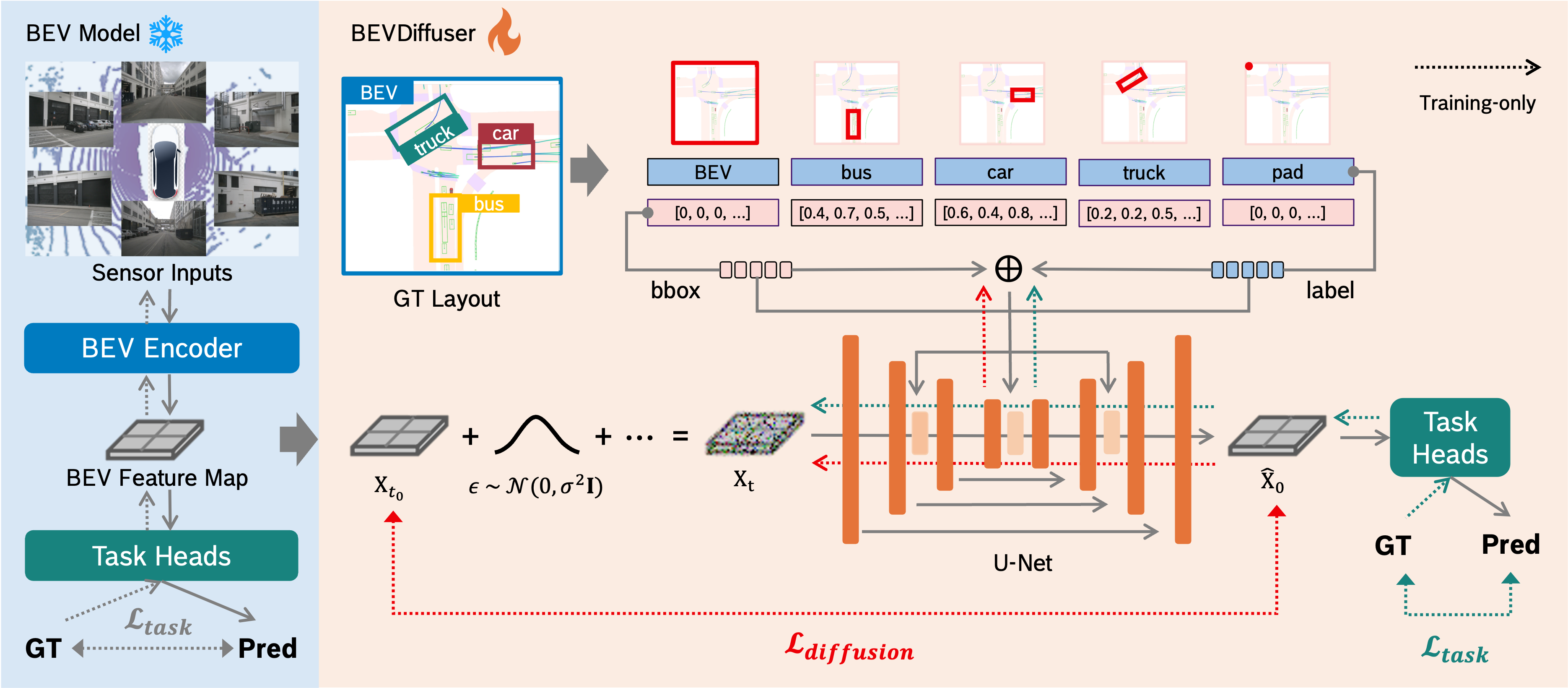}
    \caption{Left: A sketch of common BEV models that generate BEV feature maps from sensor inputs through a BEV encoder. BEV feature maps are usually optimized for downstream task performance. Right: Overview of BEVDiffuser, which consists of a U-Net that predicts the clean BEV features from the noisy ones, conditioned on the ground-truth layout. It is trained on BEV feature maps produced by BEV models with multiple steps of noise added, and is optimized using a joint loss composed of a diffusion loss and a downstream task loss. }
    \label{fig:bevdiffuser}
\end{figure*}

\section{Methodology}
\label{sec:methodology}
\subsection{Preliminary}
\textbf{BEV Model.} Though various types of BEV models have been proposed as we described in Sec.~\ref{sec:bev}, their workflow can be summarized by the sketch shown in Fig.~\ref{fig:bevdiffuser} (Left). First, a BEV encoder is usually designed to generate a BEV feature map given sensor inputs, e.g., cameras \cite{bevformer}, LiDAR \cite{52_yin2021center} or both \cite{liu2023bevfusion}. The produced BEV feature map is then fed into curated task heads to solve downstream tasks, such as 3D object detection. Due to the lack of supervision on the BEV feature map, the BEV feature map is learned indirectly by optimizing the whole model to minimize the task loss $\mathcal{L}_{task}$ and enhance the task performance.

\noindent\textbf{Diffusion Model.} Diffusion model learns to generate data from random noise $\mathcal{N}(\bm{0}, \bm{I})$ by first destroying the structure of a data distribution through gradual addition of noise to the data samples, and then learning a reverse denoising process to restore the data structure. Specifically, given a timestep $t \sim Uniform(\{1, ...,T\})$, it adds $t$-step noise to a data sample $\bm{x}_0$ to get a noisy sample $\bm{x}_t$:
\begin{equation}
    \bm{x}_t = \sqrt{\bar{\alpha_t}}\bm{x_0} + \sqrt{1-\bar{\alpha_t}}\bm{\epsilon}_t,
\end{equation}
\noindent where $\bm{\epsilon}_t \sim \mathcal{N}(\bm{0}, \bm{I})$ and $\bar{\alpha_t}=\prod_{i=1}^{t}(1-\beta_t)$ in which $\beta_t \in (0, 1)$ is a hyperparameter that controls the noise strength. Diffusion model then learns a function $f_\theta(\bm{x}_t, t)$, typically modeled by a U-Net \cite{ronneberger2015u}, to estimate the $\bm{\epsilon}_t$ by minimizing the diffusion loss:
\begin{equation}
    \mathcal{L}_{diffusion} = \mathbb{E}_{\bm{x}_0, \bm{\epsilon}_t,t}\parallel\bm{\epsilon}_t-f_\theta(\bm{x}_t, t)\parallel_2^2.
\end{equation}
\noindent After training, a new data $\bm{x}_0$ can be generated from the random noise $\bm{x}_T\sim\mathcal{N}(\bm{0}, \bm{I})$ through the iterative sampling process, which is formulated as: 
\begin{equation}
\label{eq:iter}
    \bm{x}_{t-1} = \frac{1}{\sqrt{\alpha_t}}(\bm{x}_t -\frac{1-\alpha_t}{\sqrt{1-\bar{\alpha_t}}}\bm{\epsilon}_t) + \sigma_t\bm{z},
\end{equation}
\noindent where $\bm{\epsilon}_t$ is estimated by the learned function $f_\theta(\bm{x}_t, t)$, $\bm{z}\sim\mathcal{N}(\bm{0}, \bm{I})$, and $\sigma_t$ is usually set to $\beta_t$ or scaled form of $\beta_t$.

More recently, to have a control over the denoising process and generate data of interest, conditional diffusion model with classifier-free guidance is often used because of its efficiency \cite{ho2022classifier}. In particular, a condition $y$ is fed into $f_\theta$ with a certain probability during training to get the conditional estimation of the noise $\bm{\epsilon}_t$. During sampling, the noise $\bm{\epsilon}_t$ is estimated by $(1+w)f_\theta(\bm{x}_t, t, y) - wf_\theta(\bm{x}_t, t, y=\phi)$ with the weight $w$ being set to balance the conditional and unconditional estimations.

\subsection{BEVDiffuser with Ground-Truth Guidance}

We introduce BEVDiffuser, a diffusion model denoising BEV feature maps using ground-truth guidance (see Fig.~\ref{fig:bevdiffuser} Right).  Without loss of generality, given a potentially noisy BEV feature map $\bm{x}_{t_0} (0 \leq t_0 \ll T)$ generated by the BEV encoder of any BEV models, we aim to get a denoised BEV feature map $\bm{x}_0$. Following the procedure of standard diffusion model, we learn the function $f_\theta$ to estimate the noise $\bm{\epsilon}_t$ used to form $\bm{x}_t$ under the ground-truth guidance $y$.

\noindent\textbf{Ground-Truth Guidance.} BEV feature map, as its name implies, is expected to provide a holistic top-town view of the environment that clearly presents locations and scales of objects in the environment. To get such desired BEV feature map, inspired by the layout-to-image generation task \cite{zhao2019image,zheng2023layoutdiffusion} that generates images following a specified image layout, i.e. a set of objects annotated with categories and bounding boxes, we formulate our BEV denoising problem as a layout-to-BEV generation task. Particularly, we define the BEV layout $l$ using ground-truth object annotations and condition the function $f_\theta$ on the layout $l$, namely $y=l$.

Formally, we define the BEV layout $l=\{o_0, o_1, ..., o_n\}$ to represent at most $n$ objects in the environment. Each object $o_i (1\leq i \leq n) = \{c_i, b_i\}$ is represented by its category id $c_i \in [0, \mathcal{C}+1]$ and normalized 3D bounding box $b_i \in [0,1]^{d}$. Specifically, $o_0$ is a virtual unit cube that covers the whole environment with $c_0 = 0$ . In case fewer than $n$ objects are present in the environment, we pad the layout with points $o_p$, i.e., empty objects that have no shape or appearance. We define their category id as $c_p =\mathcal{C}+1$, and the 3D bounding box $b_p$ is located at position $(0, 0, 0)$, with size, orientation and velocity are all set to $0$. 

To better fuse the BEV feature map and the layout condition, we adopt LayoutDiffusion model proposed by \cite{zheng2023layoutdiffusion} as the function $f_\theta$. Specifically, a transformer-based layout fusion module is first adopted to fuse the category and bounding box information of each object and model the relationship among them. Then the embedding of the object $o_0$ that contains the information of the entire layout is used for global conditioning. Meanwhile, the embedding of all the objects is fed into an object-aware cross attention mechanism for local conditioning. In this way, the model has better control over all the objects specified in the layout. More details can be found in supplementary materials.

\noindent\textbf{Training.} In the absence of ground-truth BEV feature map $\bm{x}_0$, we add noise $\bm{\hat{\epsilon}}_t$ to the predicted BEV  $\bm{x}_{t_0} (0 \leq t_0 \ll T)$ to get $\bm{x}_t$. In this case, we don't have access to the true noise $\bm{\epsilon}_t$, which is supposed to be added to  $\bm{x}_0$ to generate $\bm{x}_t$. As a result, instead of using $f_\theta$ to estimate the unknown $\bm{\epsilon}_t$, we propose to optimize $f_\theta$ towards $\bm{x}_0$. Since $\bm{x}_{t_0}$ is already a good estimation of $\bm{x}_0$ with bounded task errors, we first optimize $f_\theta$ towards $\bm{x}_{t_0}$ by minimizing the diffusion loss $\mathcal{L}_{diffusion}$ defined in Equation~\ref{eq:loss_diff}. To further improve the estimation accuracy, we attach task heads to consume the outputs of $f_\theta$ and generate task-specific predictions. In this way, $f_\theta$ can also be optimized through the task-specific loss $\mathcal{L}_{task}$. To sum up, we adopt the weighted sum of both losses as the overall loss $\mathcal{L}_{total}^{\scriptscriptstyle diff}$ to train $f_\theta$. Equation~\ref{eq:loss} defines the loss $\mathcal{L}_{total}^{\scriptscriptstyle diff}$ where $\lambda$ denotes a weight.

\begin{equation}
\label{eq:loss_diff}
    \mathcal{L}_{diffusion} = \mathbb{E}_{\bm{x}_{t_0}, \bm{\hat{\epsilon}}_t,t}\parallel\bm{x}_{t_0}-f_\theta(\bm{x}_t, t, y)\parallel_2^2
\end{equation}
\begin{equation}
\label{eq:loss}
    \mathcal{L}_{total}^{\scriptscriptstyle diff} = \mathcal{L}_{diffusion} + \lambda \mathcal{L}_{task}
\end{equation}

\noindent\textbf{Sampling.} We adopt classifier-free guidance in sampling process where we interpolate between conditional and unconditional outputs of $f_\theta$ to get the final estimation of $\bm{x}_0$ as Equation~\ref{eq:x0} calculates. The unconditional estimation of $\bm{x}_0$ is obtained by replacing the conditioning layout $l$ with the empty layout $l_\phi = \{o_0, o_p, ..., o_p\}$ that only contains points $o_p$. We then derive $\bm{\epsilon}_t$ from the estimated $\bm{x}_0$ by Equation~\ref{eq:epsilon} for the iterative sampling process (Equation~\ref{eq:iter}).

\begin{equation}
\label{eq:x0}
    \bm{x}_0 = (1+w)f_\theta(\bm{x}_t, t, y=l) - wf_\theta(\bm{x}_t, t, y=l_\phi)
\end{equation}
\begin{equation}
\label{eq:epsilon}
    \bm{\epsilon}_t = (\bm{x}_t - \sqrt{\bar{\alpha_t}}\bm{x}_0) / \sqrt{1-\bar{\alpha_t}}
\end{equation}

\subsection{Plug-and-Play BEVDiffuser}

BEVDiffuser can be used in a plug-and-play manner. It can be easily plugged into any BEV models during training time without changing their model architectures. During inference time, BEVDiffuser is deactivated and removed, yielding an enhanced BEV model with the same architecture to be deployed. As a result, comparing to the original BEV model, our BEVDiffuser enhanced model provides improved performance without necessitating any adaptation efforts or introducing additional computational overhead.

\begin{figure}[t!]
    \centering
    \includegraphics[width=0.47\textwidth]{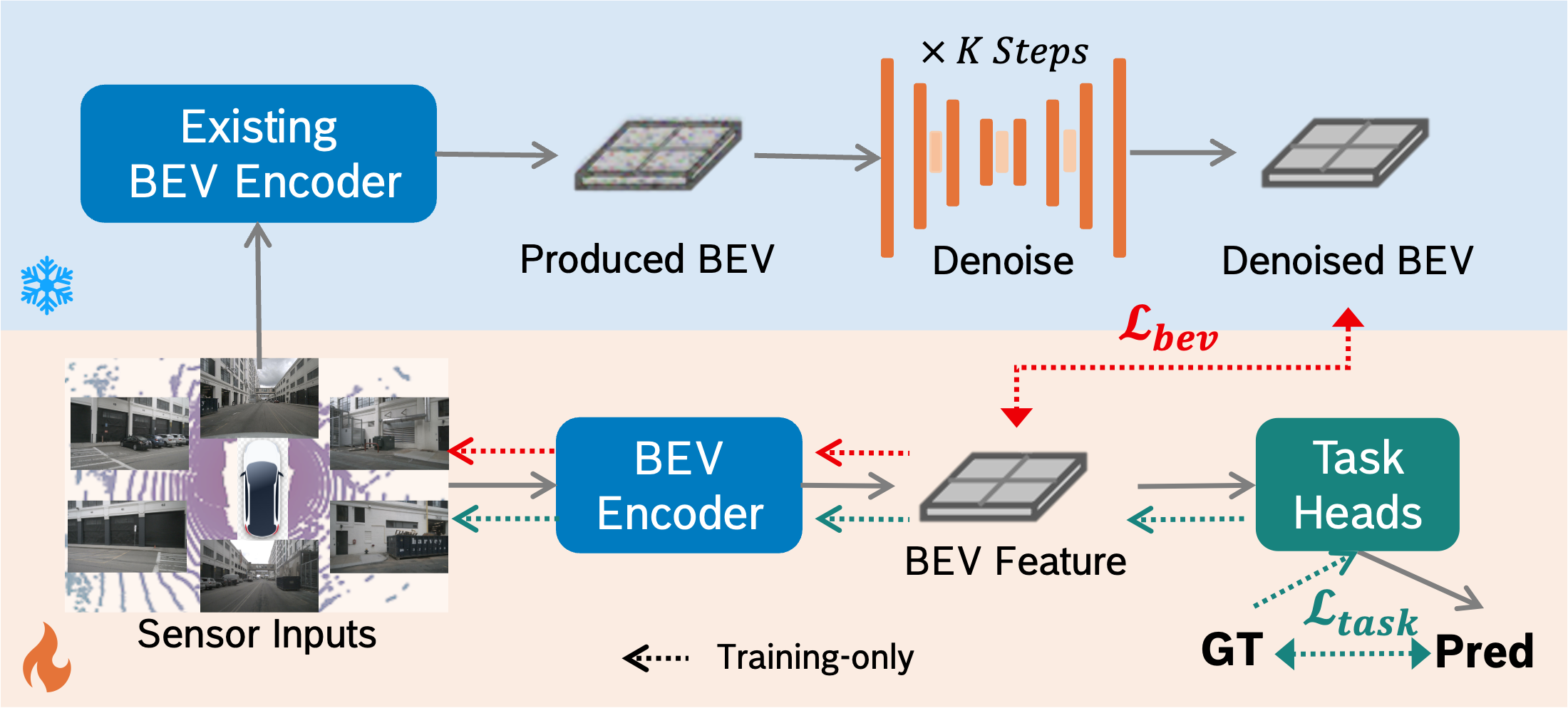}
    \caption{BEVDiffuser can be plugged into the training process of a BEV model. It denoises the BEV feature maps produced by existing BEV encoders over $K$ steps and provides the denoised BEV as supervision for BEV predictions.}
    \vspace{-20pt}
    \label{fig:plug-and-play}
\end{figure}

To be specific, as Fig.~\ref{fig:plug-and-play} depicts, given an existing BEV encoder that is originally learned with the task heads through task-specific loss $\mathcal{L}_{task}$, we denote its produced BEV feature map as $\bm{x}^{\scriptscriptstyle BEV}_K$. We adopt the trained BEVDiffuser to denoise $\bm{x}^{\scriptscriptstyle BEV}_K$ for $K$ steps and obtain the denoised BEV feature map $\bm{x}^{\scriptscriptstyle BEV}_0$. To train a new BEV model, we take $\bm{x}^{\scriptscriptstyle BEV}_0$ as a proxy ground truth of BEV and use it to supervise the new predicted BEV feature map $\bm{x}^{\scriptscriptstyle BEV}$ through loss $\mathcal{L}_{BEV}$.  $\mathcal{L}_{BEV}$ is an MSE loss defined in Equation~\ref{eq:loss_bev}. Together with the task-specific loss $\mathcal{L}_{task}$, we train the new BEV model end-to-end through the overall loss $\mathcal{L}_{total}^{\scriptscriptstyle BEV}$ shown in Equation~\ref{eq:loss_total_bev}, where $\lambda_{BEV}$ is a scaling factor.
\begin{equation}
\label{eq:loss_bev}
    \mathcal{L}_{BEV} = \mathbb{E}_{\bm{x}^{\scriptscriptstyle BEV}}\parallel\bm{x}^{\scriptscriptstyle BEV}_0-\bm{x}^{\scriptscriptstyle BEV}\parallel_2^2
\end{equation}
\begin{equation}
\label{eq:loss_total_bev}
    \mathcal{L}_{total}^{\scriptscriptstyle BEV} = \mathcal{L}_{task} + \lambda_{BEV}\mathcal{L}_{BEV}
\end{equation}
\section{Experiments}
\label{sec:experiments}
We validate BEVDiffuser on 3D object detection task, the most common task used to evaluate the effectiveness of the learned BEV feature maps \cite{bevformer,liu2023bevfusion,le2024diffuser, zou2024diffbev}. 3D object detection is critical in autonomous driving that requires both semantic and geometric understanding of the environment to identify and locate objects in 3D space. In this section, we first introduce our experimental setting in Sec.~\ref{sec:exp_setting}. In Sec.~\ref{sec:capacity}, we showcase the capacity of BEVDiffuser in denoising and generating BEV feature maps. We further demonstrate plug-and-play performance of BEVDiffuser in Sec.~\ref{sec:pp_performance} by comparing BEVDiffuser enhanced BEV models with their baseline counterparts.

\subsection{Experimental Settings}
\label{sec:exp_setting}

\textbf{Dataset.} We conduct experiments on large-scale nuScenes \cite{nuscenes} dataset. nuScenes is a well-established benchmark for autonomous driving tasks that contains 1,000 20-second driving videos, with keyframes annotated at 2 Hz. Specifically, for 3D object detection task, each keyframe provides six RGB images and a LiDAR scan covering a 360-degree field of view, as well as annotated 3D bounding boxes for objects of interest,  which are categorized by one of 10 predefined object classes. In total, the dataset contains 1.4 million annotated bounding boxes, making it well-suited for object detection task. 

\noindent\textbf{Metrics.} We adopt the official evaluation metrics provided by nuScenes detection benchmark \cite{nuscenes} to evaluate the 3D object detection performance. Specifically, mean average precision (mAP) calculates average precision by defining a true positive based on the 2D center distance between predictions and ground truth. The five true positive metrics, namely ATE, ASE, AOE, AVE, and AAE measure average translation, scale, orientation, velocity, and attribute errors,  respectively. nuScenes detection score (NDS) consolidates all the metrics into a weighted sum.

\noindent\textbf{BEV Models.} We apply BEVDiffuser to four representative and widely adopted BEV models, namely BEVFormer-tiny \cite{bevformer}, BEVFormer-base \cite{bevformer}, BEVFormerV2 \cite{bevformerv2}, and BEVFusion \cite{liu2023bevfusion}. BEVFormer and BEVFormerV2 are transformer-based methods that detect objects from only cameras, while BEVFusion adopts LSS-based method for camera inputs and then fuses camera and LiDAR features for object detection. Comparing to BEVFormer-base, BEVFormer-tiny shortens temporal dependencies and produces much smaller BEV feature maps, thereby requiring less computational cost and enabling fast development. BEVFormerV2 is a two-stage detector where a perspective head is introduced to train the image backbones and generate object proposals for the detection head. To save the computational cost, we adopt its simplest version which involves no temporal information and employs Deformable DETR \cite{zhudeformable} as the detection head.

\begin{figure}[t!]
    \centering
    \includegraphics[width=0.47\textwidth]{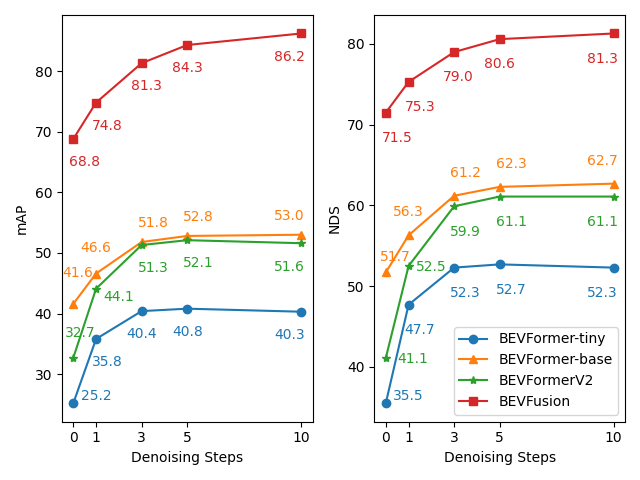}
    \caption{3D object detection performance of various BEV models on nuScenes \texttt{val} dataset (denoising steps $=0$). The performance ramps up when adopting BEVDiffuser to denoise their BEV feature maps with increasing denoising steps, indicating the powerful denoising capability of our BEVDiffuser. }
    \label{fig:bevdiffuser_plot}
\end{figure}

\subsection{Capacity of BEVDiffuser}
\label{sec:capacity}

To validate the capacity of BEVDiffuser, we train BEVDiffuser on BEV feature maps produced by each pretrained BEV model, i.e. BEVFormer-tiny, BEVFormer-base, BEVFormerV2, and BEVFusion, and we denote the trained BEVDiffuser as $\mathrm{BD}^{tiny}$,  $\mathrm{BD}^{base}$, $\mathrm{BD}^{V2}$, and $\mathrm{BD}^{fu}$, respectively. In particular, since the size of the BEV produced by  BEVFormer-base, BEVFormerV2, and BEVFusion is too large that hinders the efficient training of the diffusion models, we attach downsample and upsample layers before and after the diffusion models to reduce and restore the BEV size accordingly. Given that BEVFormer-base and BEVFormerV2 share a similar BEV feature space with BEVFormer-tiny,  we employ the trained $\mathrm{BD}^{tiny}$ as their diffusion models and only train the downsample and upsample layers to get $\mathrm{BD}^{base}$ and  $\mathrm{BD}^{V2}$. 

\begin{table*}[ht!]
\begin{center}
\begin{tabular}{l|cc|cc|ccccc}
\specialrule{0.12em}{0pt}{2pt}
Method & Mod. & BEV Size & NDS$\uparrow$ & mAP$\uparrow$ & mATE$\downarrow$ & mASE$\downarrow$ & mAOE$\downarrow$ &mAVE$\downarrow$ & mAAE$\downarrow$\\
\specialrule{0.12em}{1pt}{1.5pt}
BEVFormer-tiny \cite{bevformer} &C &$50\times50$ &35.5 &25.2 &0.898 &0.293 &0.650 &0.656 &0.216 \\
\textbf{+ BEVDiffuser} &C &$50\times50$ &\textbf{39.1} &\textbf{28.3} &\textbf{0.859} &\textbf{0.285} &\textbf{0.558} &\textbf{0.592} &\textbf{0.212} \\
\specialrule{0.01em}{1pt}{1.5pt}
BEVFormer-base \cite{bevformer} &C &$200\times200$ &51.8 &41.7 &0.673 &0.273 &0.371 &0.393 &0.198 \\
\textbf{+ BEVDiffuser} &C &$200\times200$ &\textbf{53.7} &\textbf{43.0} &\textbf{0.638} &0.274 &\textbf{0.333} &\textbf{0.355} &\textbf{0.179} \\
\specialrule{0.01em}{1pt}{1.5pt}
BEVFormerV2-base$^*$ \cite{bevformerv2} &C &$200\times200$ &41.1 &32.7 &0.768 &0.285 &0.499 &0.780 &0.195 \\
\textbf{+ BEVDiffuser} &C &$200\times200$ &\textbf{44.7} &\textbf{37.1} &\textbf{0.718} &0.286 &\textbf{0.448} &\textbf{0.740} &0.197 \\
\specialrule{0.01em}{1pt}{1.5pt}
BEVFusion$^*$ \cite{liu2023bevfusion} &LC &$180\times180$ &70.9 &67.6 &0.278 &0.253 &0.305 &0.267 &0.188 \\
\textbf{+ BEVDiffuser} &LC &$180\times180$ &\textbf{71.9} &\textbf{69.2} &\textbf{0.276} &\textbf{0.252} &\textbf{0.294} &\textbf{0.266} &\textbf{0.184} \\
\specialrule{0.12em}{1.5pt}{0pt} 
\end{tabular}
\caption{Comparison of 3D object detection performance on nuScenes \texttt{val} dataset. Our BEVDiffuser brings consistent performance improvement to existing BEV models, with notable gains in NDS and mAP.  ``Mod.'' abbreviates modality, where ``L'' and ``C'' denote LiDAR and camera, respectively. ($^*$ : model retrained under the same code base and GPU resources as its counterpart for fair comparison.) }
\label{tbl:val_performance}
\end{center}
\end{table*}

\begin{figure}[t!]
    \centering
    \includegraphics[width=0.47\textwidth]{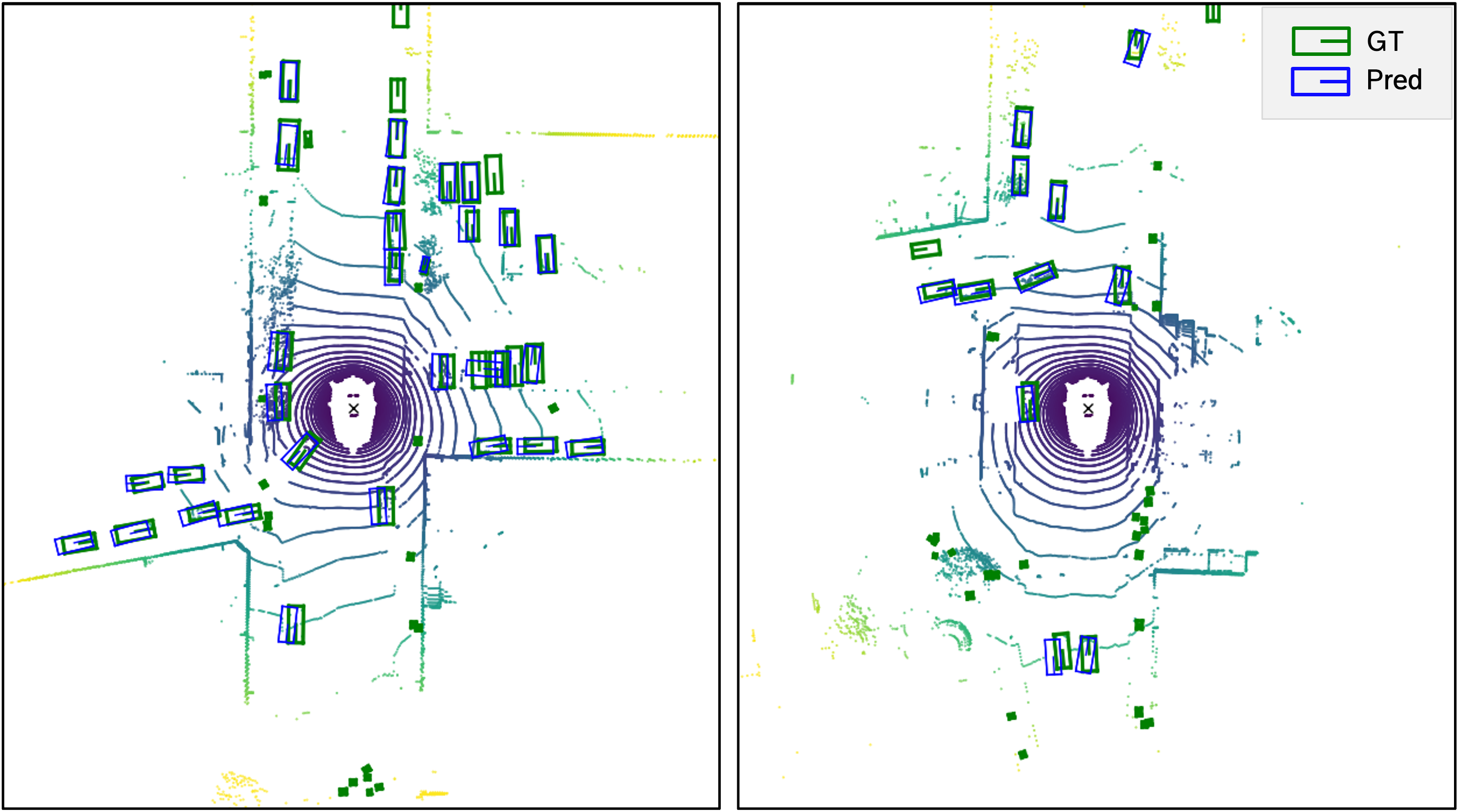}
    \caption{3D object detection visualizations of two BEV feature maps generated by our BEVDiffuser ($\mathrm{BD}^{fu}$) from random noise. The alignment between predictions and ground truth demonstrates that BEVDiffuser has strong controllable generation capability.  }
    \label{fig:gen_vis}
\end{figure}

\noindent\textbf{BEV Denoising Capability.} We use the trained BEVDiffuser to denoise the BEV feature maps from each BEV model and assess their 3D object detection performance using the denoised features. Fig.~\ref{fig:bevdiffuser_plot} reports the mAP and NDS achieved on the nuScenes \texttt{val} dataset. Noticeably, the detection performance of all BEV models has been significantly improved after the BEV feature maps are denoised. The performance grows sharply when the number of denoising steps gradually increases to $5$, demonstrating the powerful denoising capability of BEVDiffuser. After denoising the BEV feaure maps for $5$ steps, the performance growth slows down, which is expected since less noise remains. This observation further confirms BEVDiffuser's efficiency in denoising BEV feature maps.

\noindent\textbf{BEV Generation Capability.} BEVDiffuser as a conditional diffusion model is also able to generate a BEV feature map from a conditioning layout. To evaluate its BEV generation capability, we use the trained BEVDiffuser ($\mathrm{BD}^{fu}$) to generate BEV feature maps from random noise $\mathcal{N}(\bm{0}, \bm{I})$, conditioned on the ground-truth layout built from nuScenes \texttt{mini-val} dataset. To speed up the generation process, we adopt DDIM scheduler \cite{songdenoising} to skip steps in denoising process. In practice, we run $50$ denoising steps to generate the BEV feature maps. We further decode the generated BEV feature maps using the pretrained detection head from BEVFusion and achieve $41.1\%$ NDS and $36.7\%$ mAP for detection on nuScenes \texttt{mini-val} dataset. We visualize the detection results from the LiDAR top view in Fig.~\ref{fig:gen_vis}. As shown in the figure, the predictions using the generated BEV feature maps align well with the ground truth, showing the strong controllable generation capability of BEVDiffuser. This capability makes BEVDiffuser even promising in augmenting data for corner cases and developing driving world model \cite{wang2024driving, yang2024generalized, gao2024vista} in the BEV feature space, which we leave for future research.

\begin{table}[b!]
    \centering
    \resizebox{1.0\linewidth}{!}{
     \begin{tabular}{l|cccc}
        \toprule
        Model {\footnotesize(\textbf{+ BEVDiffuser}) } & Mod. & BEV Size & \# Params & FPS \\
        \midrule
        BEVFormer-tiny &  C & $50\times50$ &  33.6 M & 6.0 \\
        \midrule
        BEVFormer-base & C  &   $200\times200$ & 69.1 M  & 2.7 \\
        \midrule
        BEVFormerV2-base & C  & $200\times200$& 56.3 M  & 3.2 \\
        \midrule     
        BEVFusion & LC  & $180\times180$ & 40.8 M & 2.9$^\dag$ \\
        \bottomrule
    \end{tabular}
    }
    \caption{Computational efficiency tested on 1 A100 GPU. Plugging in BEVDiffuser doesn't change the network architecture and therefore maintain the same computational efficiency as the baselines. ($\dag$: tested on official MMCV implementation) } 
    \label{tab:comp_efficiency}
\end{table}

\subsection{Plug-and-Play Performance of BEVDiffuser}
\label{sec:pp_performance}

\begin{table*}[ht!]
\begin{center}
\begin{tabular}{l|cccccccccc}
\specialrule{0.12em}{0pt}{2pt}
Method &\rot{\makecell[l]{Constr. Veh.\\{\small(1.0\%)}}} &\rot{\makecell[l]{Bus\\{\small(1.0\%)}}} &\rot{\makecell[l]{Motorcycle\\{\small(1.2\%)}}} &\rot{\makecell[l]{Bicycle\\{\small(1.3\%)}}} &\rot{\makecell[l]{Trailer\\{\small(1.7\%)}}} &\rot{\makecell[l]{Truck\\{\small(6.5\%)}}} &\rot{\makecell[l]{Traf. Cone\\{\small(10.2\%)}}} &\rot{\makecell[l]{Barrier\\{\small(15.9\%)}}} &\rot{\makecell[l]{Pedestrian\\{\small(18.0\%)}}} &\rot{\makecell[l]{Car\\{\small(43.1\%)}}} \\
\specialrule{0.12em}{1pt}{1.5pt}
BEVFormer-tiny \cite{bevformer} &5.8 &23.4 &21.4 &20.3 &6.6 &19.2 &38.4 &37.9 &33.2 &45.7 \\
\textbf{+ BEVDiffuser} &\textbf{7.2} &\textbf{30.3} &\textbf{26.9} &\textbf{24.0} &\textbf{8.2} &\textbf{22.8} &\textbf{40.7} &\textbf{40.0} &\textbf{34.8} &\textbf{48.1} \\
\specialrule{0.01em}{1pt}{1.5pt}
BEVFormer-base \cite{bevformer}  &12.9 &44.5 &43.0 &39.8 &17.2 &37.0 &58.5 &52.6 &49.4 &61.9 \\
\textbf{+ BEVDiffuser} &\textbf{13.5} &\textbf{47.1} &\textbf{44.8} &\textbf{41.7} &\textbf{18.0} &\textbf{37.2} &\textbf{59.6} &\textbf{55.6} &\textbf{50.3} &61.8\\
\specialrule{0.01em}{1pt}{1.5pt}
BEVFormerV2-base$^*$ \cite{bevformerv2}  &3.4 &33.7 &29.8 &25.6 &7.5 &26.5 &52.4 &50.1 &42.8 &55.5 \\
\textbf{+ BEVDiffuser} &\textbf{6.4} &\textbf{41.8} &\textbf{35.1} &\textbf{30.1} &\textbf{11.8} &\textbf{32.0} &\textbf{55.5} &\textbf{54.5} &\textbf{45.0} &\textbf{58.8}\\
\specialrule{0.01em}{1pt}{1.5pt}
BEVFusion$^*$ \cite{liu2023bevfusion}  &29.9 &74.9 &75.3 &60.4 &46.7 &62.4 &79.3 &70.2 &88.1 &89.3\\
\textbf{+ BEVDiffuser} &\textbf{30.9} &\textbf{76.6} &\textbf{76.9} &\textbf{63.3} &\textbf{48.4} &\textbf{65.2} &\textbf{79.9} &\textbf{72.9} &\textbf{88.3} &\textbf{89.5} \\
\specialrule{0.12em}{1.5pt}{0pt} 
\end{tabular}
\caption{Per-class object detection results (mAP) on nuScenes \texttt{val} dataset. Note that object classes are sorted based on the percentage of their occurrences in the dataset (shown under the class names). BEVDiffuser exhibits overall improvements across all classes, with more significant gains on long-tail objects that appears only 1-2$\%$ in the dataset, such as \textit{construction vehicle} and \textit{bus}. }
\label{tbl:class_performance}
\end{center}
\end{table*}

\begin{table*}[ht!]
\begin{center}
\begin{tabular}{lcccccccccccc}
\specialrule{0.12em}{0pt}{2pt}
& &\multicolumn{2}{c}{Sunny}&& \multicolumn{2}{c}{Rainy}&& \multicolumn{2}{c}{Day}&& \multicolumn{2}{c}{Night}\\
\cline{3-4}\cline{6-7}\cline{9-10}\cline{12-13}
Method & Mod. & NDS$\uparrow$ & mAP$\uparrow$ && NDS$\uparrow$ & mAP$\uparrow$ &&NDS$\uparrow$ & mAP$\uparrow$ &&NDS$\uparrow$ & mAP$\uparrow$\\
\specialrule{0.12em}{1pt}{1.5pt}
BEVFormer-tiny \cite{bevformer} &C &34.9 &25.0 &&37.7 &26.9 &&35.8 &25.6 &&18.1 &9.5 \\
\textbf{+ BEVDiffuser} &C &\textbf{38.4} &\textbf{28.0} &&\textbf{42.2} &\textbf{30.1} &&\textbf{39.4} &\textbf{28.7} &&\textbf{19.5} &\textbf{11.4} \\
\specialrule{0.01em}{1pt}{1.5pt}
BEVFormer-base \cite{bevformer} &C &50.9 &41.1 &&55.2 &43.8 &&52.0 &41.9 &&28.4 &21.1 \\
\textbf{+ BEVDiffuser} &C &\textbf{52.9} &\textbf{42.4} &&\textbf{56.5} &\textbf{45.2} &&\textbf{54.0} &\textbf{43.3} &&\textbf{30.4} &\textbf{22.6} \\
\specialrule{0.01em}{1pt}{1.5pt}
BEVFormerV2-base$^*$ \cite{bevformerv2} &C &40.2 &32.7 &&44.8 &31.7 &&41.5 &33.3 &&18.6 &11.4 \\
\textbf{+ BEVDiffuser} &C &\textbf{43.4} &\textbf{36.4} &&\textbf{49.4} &\textbf{38.8} &&\textbf{45.1} &\textbf{37.7} &&\textbf{21.2} &\textbf{14.7} \\
\specialrule{0.01em}{1pt}{1.5pt}
BEVFusion$^*$ \cite{liu2023bevfusion} &LC &70.5 &67.0 &&72.8 &69.4 &&71.1 &67.7 &&44.0 &39.9 \\
\textbf{+ BEVDiffuser} &LC &\textbf{71.5} &\textbf{68.9} &&\textbf{72.9} &\textbf{69.6} &&\textbf{72.0} &\textbf{69.3} &&\textbf{45.2} &\textbf{41.3} \\
\specialrule{0.12em}{1.5pt}{0pt} 
\end{tabular}
\caption{Object detection performance on nuScenes \texttt{val} dataset under different weather and lighting conditions. BEVDiffuser consistently improves upon its baseline counterparts in all scenarios across all metrics. In particular, we observe significant improvements at night scenarios, when poor lighting conditions pose a significant challenge for camera-based perception.  }
\vspace{-15pt}
\label{tbl:weather_performance}
\end{center}
\end{table*}

BEVDiffuser can be a plug-and-play module for state-of-the-art BEV models without any bells and whistles. Here, we plug the trained BEVDiffuser into the training process of BEVFormer-tiny, BEVFormer-base, BEVFormerV2, and BEVFusion, respectively. We use BEVDiffuser to denoise the existing BEV feature maps for $5$ steps and train new BEV models from scratch under the supervision of the denoised feature maps to get the BEVDiffuser enhanced models. We compare the BEVDiffuser enhanced models with their baseline counterparts to assess the plug-and-play performance of the BEVDiffuser.

\noindent\textbf{3D Object Detection Comparison.} We report the 3D object detection performance of all models achieved on nuScenes \texttt{val} dataset in Tab.~\ref{tbl:val_performance}.  As shown in the table, our BEVDiffuser enhanced models consistently outperform their baseline counterparts across almost all the metrics, especially in NDS and mAP. Notably, our BEVDiffuser enhanced BEVFormer-tiny raises NDS and mAP by $10.1\%$ and $12.3\%$ respectively. Similarly, BEVDiffuser boosts BEVFormerV2 by achieving $8.8\%$  and $13.5\%$ improvement in NDS and mAP. For more complex BEV models, i.e. BEVFormer-base and BEVFusion, where their BEV feature maps have been well learned as shown by their outstanding object detection performance, our BEVDiffuser continues to effectively denoise their BEV feature maps, guide their training process, and consistently improve the performance.

It is worth highlighting that BEVDiffuser brings performance enhancement to BEV models at no cost of any additional adaptation efforts or computational overhead. As a training-only plug-in, BEVDiffuser is removed at deployment, leaving an enhanced BEV model with the architecture unchanged, which is then used for testing. As a result, our BEVDiffuser enhanced models share the same network size and latency as their baseline counterparts which are summarized in Tab.~\ref{tab:comp_efficiency}. Unlike previous work \cite{le2024diffuser, zou2024diffbev} that need to pass their integrated diffusion models multiple times to denoise the BEV feature maps on-the-fly, our method is more flexible and superior in latency-critical applications like autonomous driving.

\begin{figure*}[ht!]
    \centering
    \includegraphics[width=1.0\textwidth]{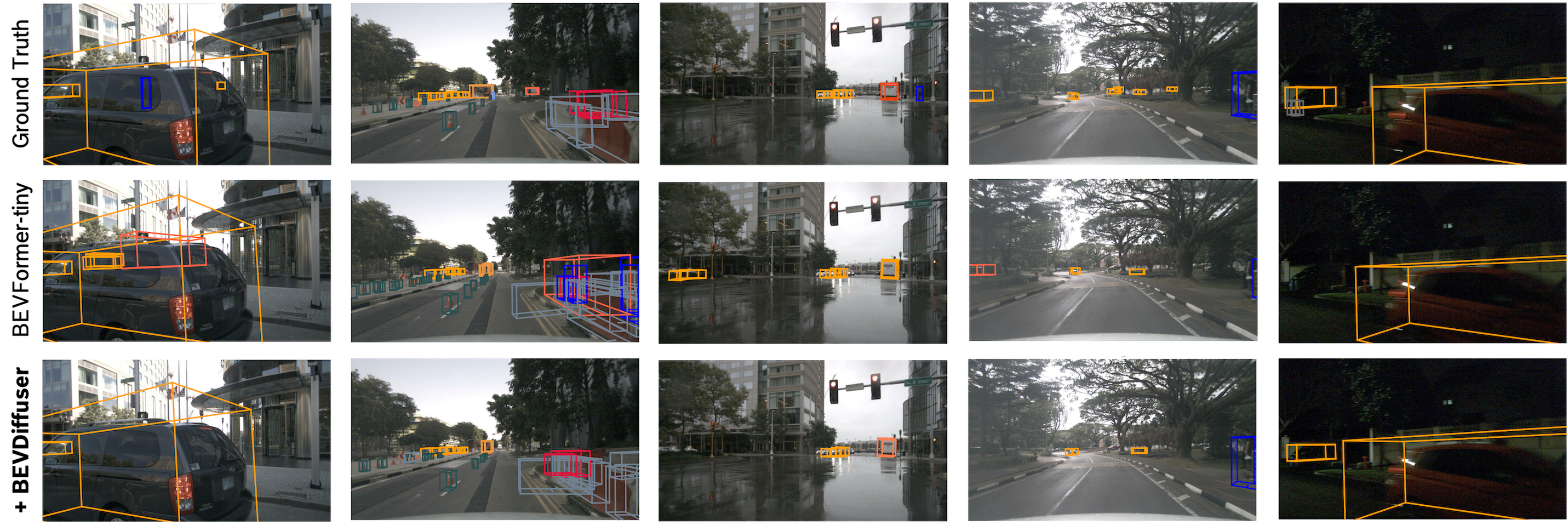}
    \caption{Visualization results of BEVDiffuser enhanced BEVFormer-tiny on nuScenes \texttt{val} dataset. Compared to the baseline BEVFormer-tiny, BEVDiffuser helps to reduce hallucinations (first three columns) and detect safety-critical objects (last two columns).  }
    \label{fig:vis}
\end{figure*}

\noindent\textbf{Performance on Long-tail Objects.} BEV feature maps, optimized only for downstream task performance, tend to misclassify and overlook underrepresented objects. As illustrated in Tab.~\ref{tbl:class_performance} where per-class object detection results are presented, all baseline models are more effective at detecting the predominant object \textit{car}, compared to long-tail objects like \textit{construction vehicle} and \textit{bus}, which appear only 1-2\% of the time. In contrast, BEVDiffuser denoises BEV feature maps using ground-truth layout as guidance that captures the joint distributions of all objects. As a result, BEVDiffuser exhibits overall improvements across all classes as demonstrated in  Tab.~\ref{tbl:class_performance}. Notably, it achieves more substantial gains for long-tail objects. For example, BEVDiffuser improves BEVFormer-tiny's detection of the long-tail objects, \textit{construction vehicle} and \textit{bus}, with mAP enhancement of $24.1\%$ and $29.5\%$, respectively. BEVDiffuser enhanced BEVFormerV2 also increases mAP by $88.2\%$ and $23.4\%$ for detecting \textit{construction vehicle} and \textit{bus}. The remarkable improvements in long-tail object detection emphasize the enhanced BEV feature maps learned by BEVDiffuser, showing its effectiveness in BEV denoising process.

\noindent\textbf{Robustness Analysis.} We analyze the robustness of the BEVDiffuser under different weather and lighting conditions. From Tab.~\ref{tbl:weather_performance}, BEVDiffuser consistently improves its baseline counterparts for both sunny and rainy, day and night scenarios. Specifically, while poor lighting condition at night poses significant challenge for camera-based perception, BEVDiffuser achieves $20.0\%$ and $28.9\%$ mAP improvements over the baseline BEVFormer-tiny and BEVFormerV2, respectively. In addition, on sunny days, BEVDiffuser also compensates for camera noise caused by overexposure, leading to improved detection performance. BEVFusion, which enhances robustness by using multi-modal sensors, i.e camera and LiDAR, still benefits from BEVDiffuser in challenging weather and lighting conditions. The notable improvements across all scenarios highlight the enhanced robustness delivered by BEVDiffuser.

\noindent\textbf{Qualitative Results.} Fig.~\ref{fig:vis} depicts how BEVDiffuser improves the 3D object detection performance. We show the ground-truth and the predicted 3D bounding boxes on camera images for comparison. As shown in the first three columns, BEVDiffuser reduces hallucinations generated by the baseline model, BEVFormer-tiny. Taking the second column as an example, BEVFormer-tiny mistakenly detects pedestrians nearby, as indicated by the blue bounding boxes. In comparison, our BEVDiffuser enhanced model effectively resolves such false positive detections. Moreover, BEVDiffuser also helps to minimize false negative detections. As the last two columns demonstrate, our BEVDiffuser enhanced model successfully detects the pedestrian in front of the autonomous vehicle and the car crossing the road, both of which are overlooked by the baseline model but are crucial for ensuring the autonomous vehicle's safe operation. Overall, BEVDiffuser aligns the detections more closely with the ground truth, highlighting its effectiveness in enhancing the quality of the BEV feature maps. We present more qualitative results in supplementary materials.

\section{Conclusion and Future Work}
\label{sec:conclusion}

In this work, we present BEVDiffuser, a novel diffusion model that denoises BEV feature maps using ground-truth guidance. BEVDiffuser consists of a U-Net model trained on BEV feature maps generated by existing BEV models. The U-Net model predicts clean BEV feature maps conditioned on the ground-truth object layout, which then derives the denoising process. BEVDiffuser can be used as a training-only plug-and-play module to enhance the existing BEV models by providing denoised BEV feature maps as additional supervision to BEV predictions. Extensive experiments on challenging nuScenes dataset demonstrate BEVDiffuser's exceptional denoising and generation capabilities, resulting in significant improvements to existing BEV models, without the need for architectural changes or additional computational overhead. Moreover, results on long-tail object detection and under challenging weather and lighting conditions further confirm the efficacy of BEVDiffuser in improving the BEV quality. In future work, we plan to investigate potential applications of BEVDiffuser for other autonomous driving tasks, such as motion prediction and data augmentation for corner cases.

{
    \small
    \bibliographystyle{ieeenat_fullname}
    \bibliography{main}
}

\clearpage
\setcounter{page}{1}
\maketitlesupplementary

\section{Model Architecture}
We follow Latent Diffusion Models (LDMs) \cite{rombach2022high} to build a conditional diffusion model as our BEVDiffuser by augmenting the U-Net with cross-attention layers. The cross-attention operation is defined in Equation~\ref{eq:attn}, where $W_*$ represents learnable projection matrices unless otherwise specified, $\varphi_i(\bm{x}_t)$ denotes the intermediate embedding of $\bm{x}_t$ from the $i$-th layer of the U-Net, and $\tau_\theta(y)$ indicates the embedding of the condition $y$. 
\begin{equation}
\label{eq:attn}
\begin{gathered} 
cross\mbox{-}attn(Q, K, V) = softmax(\frac{QK^T}{\sqrt{d}}) \cdot V  \\
Q = \varphi_i(\bm{x}_t)W_Q^i,\, K= \tau_\theta(y)W_K^i,\, V=\tau_\theta(y)W_V^i 
\end{gathered}
\end{equation}

To better fuse the BEV feature map $\bm{x}_t$ and the layout condition $y=l$ and have more control over all the objects specified in the layout, we adopt the global conditioning and the object-aware local conditioning mechanism proposed by \cite{zheng2023layoutdiffusion}. Specifically, we first use a transformer-based layout fusion module $LFM$ as $\tau_\theta$ to get a self-attended embedding $o'_i$ for each object $o_i$ as shown in Equation~\ref{eq:lfm}. In this way, $o'_0$ contains the information of the entire layout and is then added to $\bm{x}_t$ for global conditioning, i.e., $\bm{x}_t' = \bm{x}_t + o'_0W_o $. Meanwhile, the embedding of all the objects $l'=\{o'_i\}_{i=0}^{n}$ is used to construct the key $K_l$ and the value $V_l$ of the layout for object-aware local conditioning. We adopt convolutional operations for the construction as shown by Equation~\ref{eq:qkv_l}. Similarly, we construct the query, key and value of the BEV feature as Equation~\ref{eq:qkv_x} shows. To align the BEV feature with the layout, we divide the BEV feature map $\bm{x}_t$ equally into $k\times k$ bounding boxes, denoted by $\{b_x\}_{1}^{k\times k}$. We encode the bounding boxes from both BEV feature and layout, i.e., $b_x$ and $b_l$, into the same embedding space using the shared weights $W_b$ and $W_p$, and get the positional embedding $P_x$ and $P_l$ for the BEV feature and the layout, respectively (see Equation~\ref{eq:pos}). $P_x$ and $P_l$ are utilized to generate the fused query, key and value by combining the BEV feature and the layout for the cross-attention operation, as formulated in Equation~\ref{eq:qkv}. $[\ \cdot\ ] $ represents the concatenation operation.

\begin{equation}
\label{eq:lfm}
\begin{aligned}
 l' = \{o'_i\}_{i=0}^{n} &= LFM(\{o_i\}_{i=0}^{n}) \\
 &= self\mbox{-}attn(\{c_iW_c + b_iW_b\}_{i=0}^{n})
\end{aligned}
\end{equation}

\begin{equation}
\label{eq:qkv_l}
K_l,\, V_l = conv_{w_l}(l')
\end{equation}

\begin{equation}
\label{eq:qkv_x}
Q_x,\, K_x,\, V_x = conv_{w_x}(\varphi_i(\bm{x}_t'))
\end{equation}

\begin{equation}
\label{eq:pos}
P_x = b_xW_bW_p, \quad P_l = b_lW_bW_p
\end{equation}

\begin{equation}
\label{eq:qkv}
Q = 
\begin{bmatrix}
Q_x \\
P_x
\end{bmatrix},
\,K =
\begin{bmatrix}
K_x& K_l \\
P_x& P_l
\end{bmatrix},
\,V =
\begin{bmatrix}
V_x& V_l 
\end{bmatrix}
\end{equation}

\begin{figure}[t!]
    \centering
     \begin{subfigure}[b]{0.22\textwidth}
         \centering
         \includegraphics[width=\textwidth]{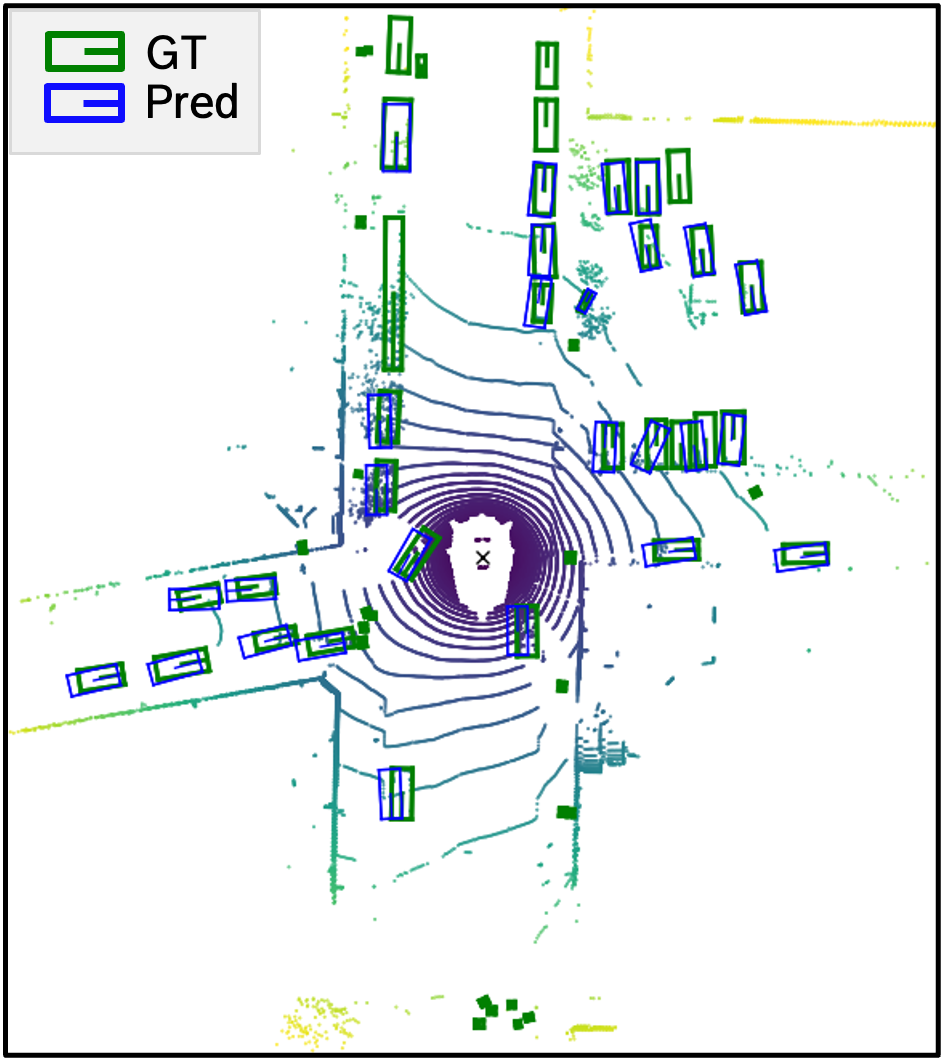}
         \caption{Existing layout.}
         \label{fig:ctrl_gen_a}
     \end{subfigure}
     \begin{subfigure}[b]{0.22\textwidth}
         \centering
         \includegraphics[width=\textwidth]{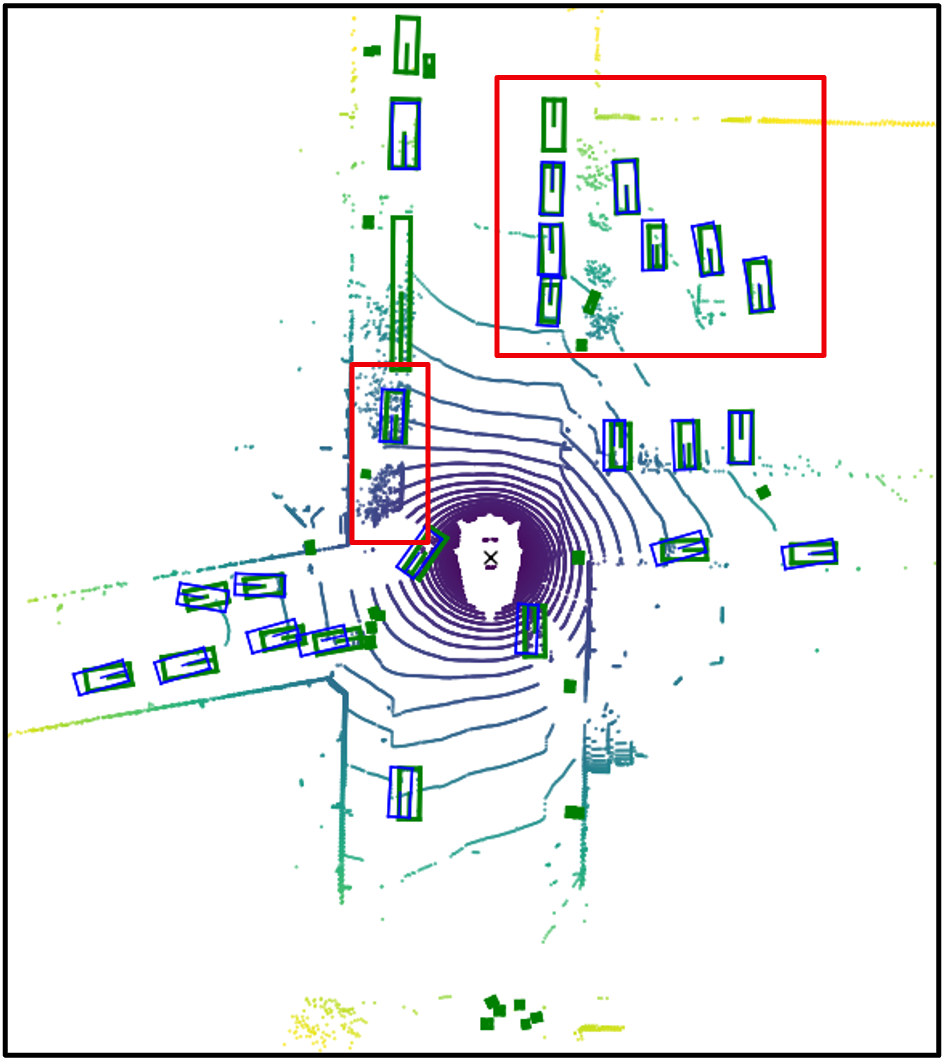}
         \caption{Objects removed.}
         \label{fig:ctrl_gen_b}
     \end{subfigure} 
     \begin{subfigure}[b]{0.22\textwidth}
         \centering
         \includegraphics[width=\textwidth]{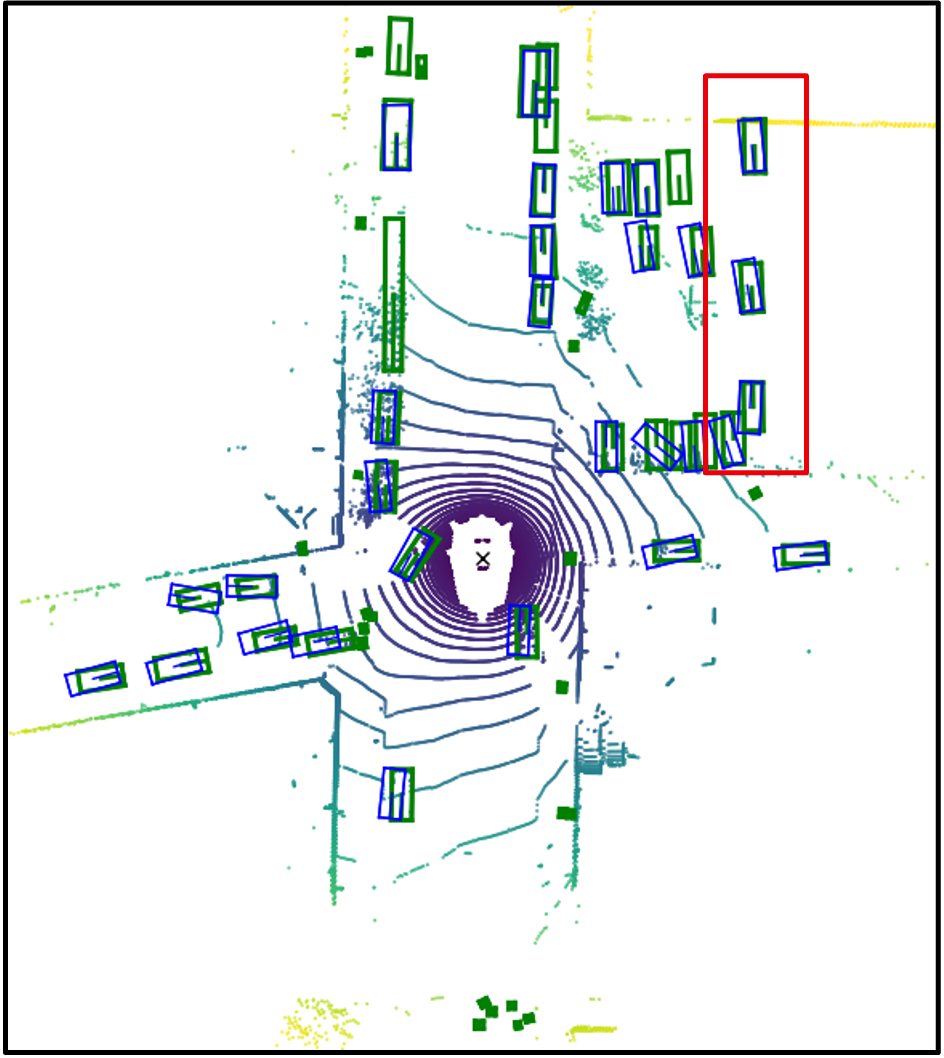}
         \caption{Objects added.}
         \label{fig:ctrl_gen_c}
     \end{subfigure}
     \begin{subfigure}[b]{0.22\textwidth}
         \centering
         \includegraphics[width=\textwidth]{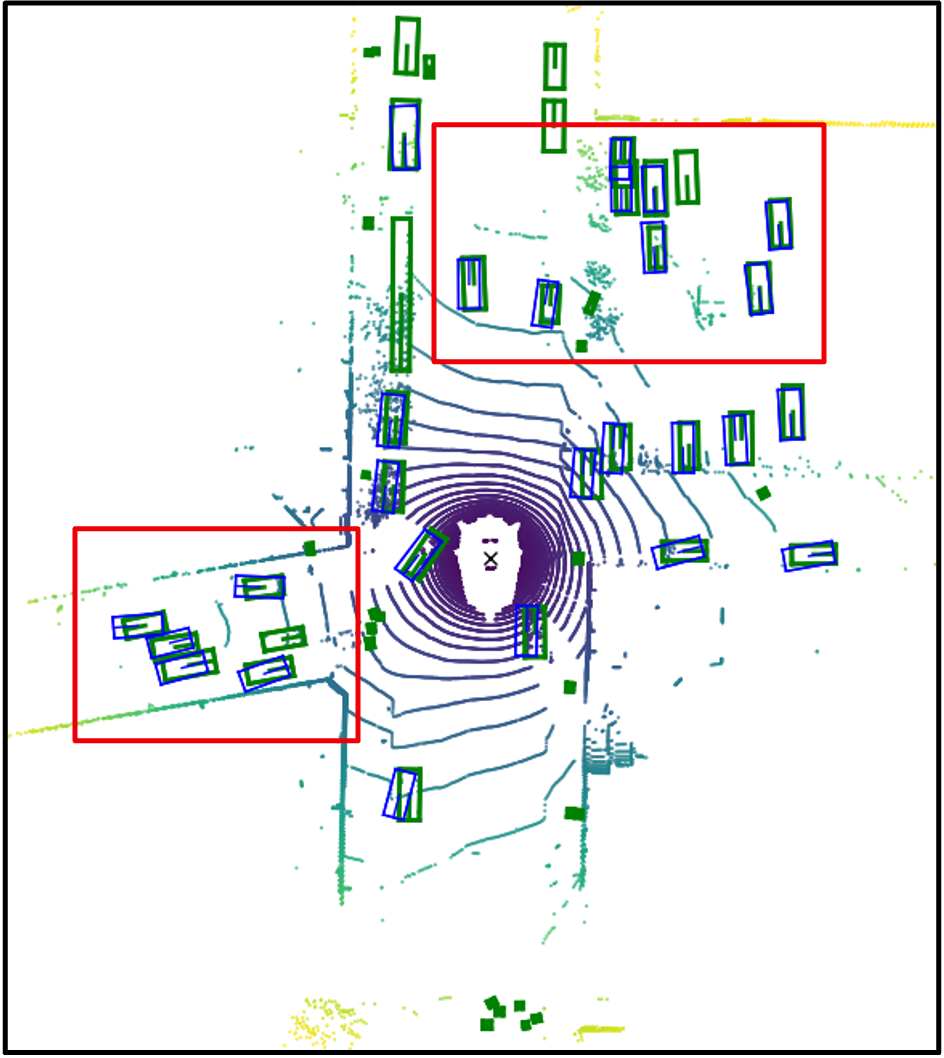}
         \caption{Objects repositioned.}
         \label{fig:ctrl_gen_d}
     \end{subfigure} 
    \caption{BEV feature maps generated by our BEVDiffuser ($\mathrm{BD}^{fu}$) from pure noise, conditioned on user-defined layouts. We modify an existing layout (a) from nuScenes \texttt{mini-val} dataset by randomly removing (b), adding (c), and repositioning (d) some objects, as highlighted by the red boxes. BEVDiffuser generates accurate BEV feature maps, enabling the detection head to produce predictions that closely align with the ground truth. }
    \label{fig:ctrl_gen}
\end{figure}

\begin{figure*}[ht!]
    \centering
    \includegraphics[width=1.0\textwidth]{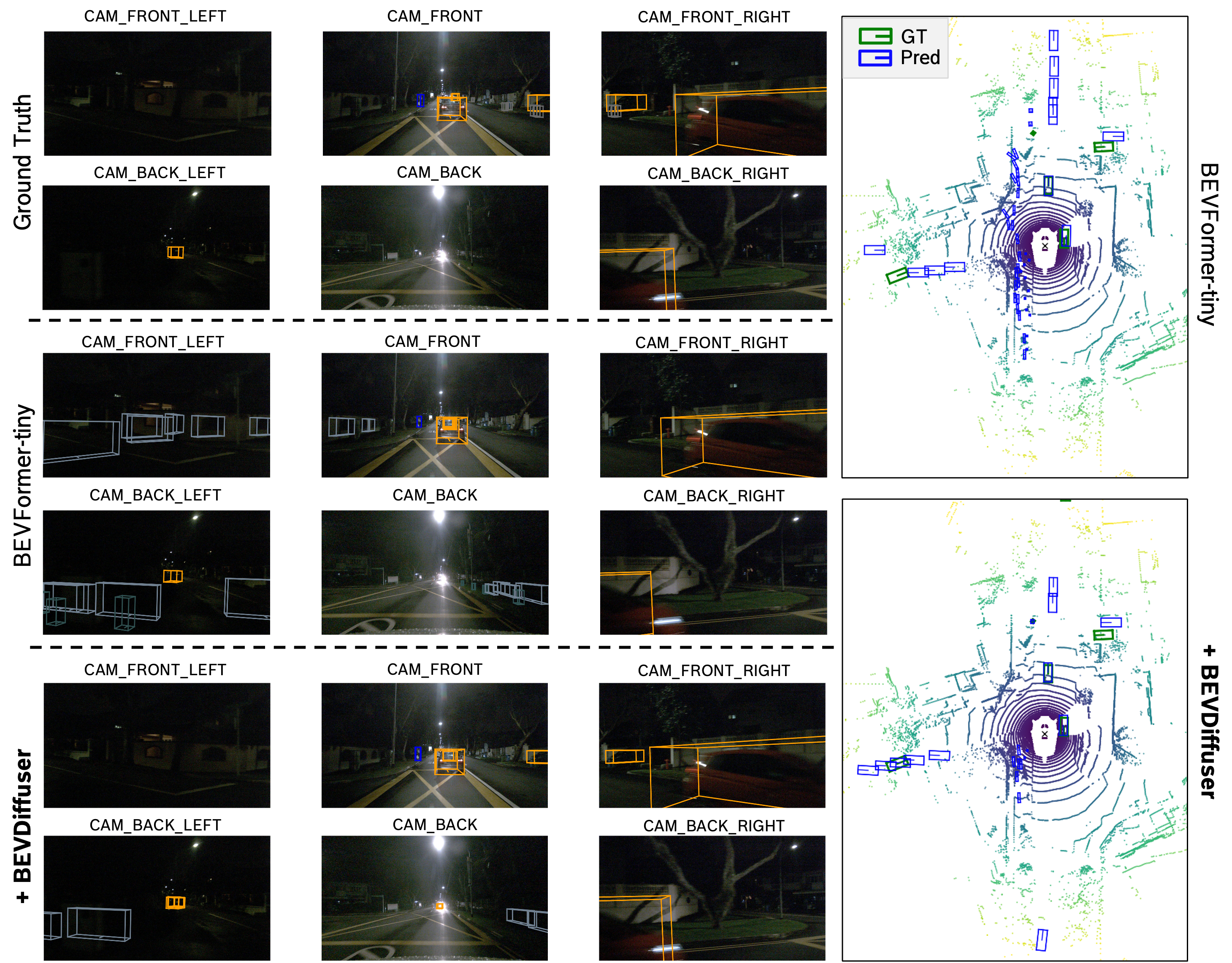}
    \caption{Visualization results of our BEVDiffuser enhanced BEVFormer-tiny on nuScenes \texttt{val} dataset. As shown in \texttt{CAM\_FRONT} and \texttt{CAM\_FRONT\_RIGHT}, BEVDiffuser helps BEVFormer-tiny to detect the car intending to cross the road under the challenging lighting condition. Moreover, BEVDiffuser also helps to reduce hallucinations generated by BEVFormer-tiny, especially on \texttt{CAM\_FRONT\_LEFT}. }
    \label{fig:vis_tiny}
\end{figure*}

\section{Implementation Details}
Our implementation is built upon the official BEVFormer implementation \footnote{\url{https://github.com/fundamentalvision/BEVFormer}} and the MMCV implementation of the BEVFusion \footnote{\url{https://github.com/open-mmlab/mmdetection3d/tree/main/projects/BEVFusion}}. The hyperparameter $\lambda$ and $\lambda_{BEV}$ are empirically tuned based on the scale of the loss. 
Specifically, we configure $\lambda$ and $\lambda_{BEV}$ as follows: for BEVFormer-tiny and BEVFormer-base, $\lambda = 0.1$ and $\lambda_{BEV} = 100$; for BEVFormerV2, $\lambda = 0.05$ and $\lambda_{BEV} = 100$; and for BEVFusion, $\lambda = 0.2$ and $\lambda_{BEV} = 20$.

\section{Ablation Study}

We conduct an ablation study on BEVDiffuser ($\mathrm{BD}^{tiny}$) to validate our design choices of layout conditioning and optimization objective, i.e. optimizing towards $x_{t_0}$ with the task loss. Note that to optimize towards $\hat{\epsilon}_{t}$, we are not able to attach the task head or use the task loss. As shown in Tab.~\ref{tbl:ablation}, without the task loss, whether we optimize towards $x_{t_0}$ or $\hat{\epsilon}_{t}$, the denoising capability we obtained is quite limited, demonstrating that the task loss is critical to guarantee the denoising performance. Similarly, our layout conditioning also contributes to the superior denoising capability of BEVDiffuser, as evidenced by the inferior performance of the unconditional model.

\begin{table}[ht!]
\setlength{\tabcolsep}{4.8pt}
\small
\begin{center}
\begin{tabular}{lccccc}
&& \multicolumn{4}{c}{\# denoising steps}\\
\cline{3-6}
 Method & obj. & 1 & 3 & 5 & 10\\
\specialrule{0.12em}{0pt}{1pt}
\textbf{Ours} &$x_{t_0}$ & \textbf{35.8/47.7} &\textbf{40.4/52.3} &\textbf{40.8/52.7} & \textbf{40.3/52.3}\\
\hline
\multirow{2}{*}{$-$task}& $x_{t_0}$ &24.5/34.7 &23.1/32.8 &21.7/31.0 &17.4/26.1 \\
&$\hat{\epsilon}_{t}$ &25.2/35.5 &25.2/35.5 &25.2/35.5 &25.2/35.5 \\
\hline
$-$cond. &$x_{t_0}$ & 25.4/35.4 & 25.3/35.3 & 25.1/35.0 & 24.7/34.6\\
\specialrule{0.12em}{0pt}{0pt}
\end{tabular}
\caption{Ablation study. mAP/NDS achieved by the variants of BEVDiffuser ($\mathrm{BD}^{tiny}$) with increasing denoising steps (1$\rightarrow$10). Results validate that both the task loss and the layout conditioning contribute to the superior denoising capability of BEVDiffuser. }
\label{tbl:ablation}
\end{center}
\end{table}

\begin{figure*}[ht!]
    \centering
    \includegraphics[width=1.0\textwidth]{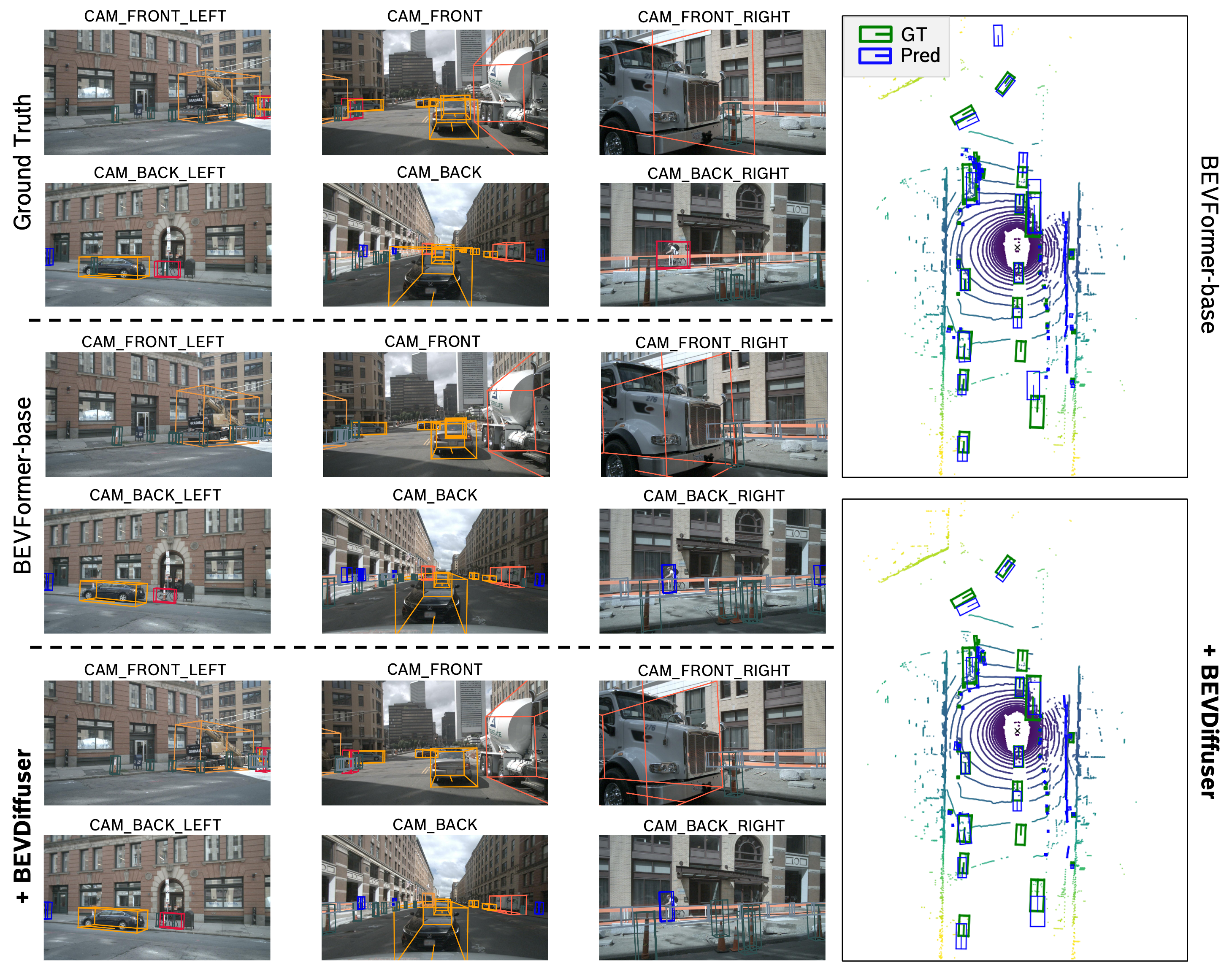}
    \caption{Visualization results of our BEVDiffuser enhanced BEVFormer-base on nuScenes \texttt{val} dataset. While BEVFormer-base shows good performance in the crowded environment, BEVDiffuser enhances its performance further, such as by detecting a human riding a bicycle in front of the autonomous vehicle, as indicated by the red bounding box in \texttt{CAM\_FRONT} and \texttt{CAM\_FRONT\_LEFT}.}
    \label{fig:vis_base}
\end{figure*}

\section{Additional Qualitative Results}
\subsection{Controllable BEV Generation}

We present user-defined layout-conditioned BEV generation in Fig.~\ref{fig:ctrl_gen}. We modify an existing layout by randomly removing, adding, or repositioning some objects, and then condition the BEVDiffuser on the modified layouts to generate BEV feature maps. As shown in Fig.~\ref{fig:ctrl_gen},  BEVDiffuser is able to produce BEV feature maps that enable accurate object detection in alignment with the specified layouts, demonstrating its strong controllable generation capability. This capability facilitates easy adjustments to object presence and positioning in the BEV feature space, paving the way for large-scale data collection and driving world model development to advance autonomous driving.

\subsection{3D Object Detection}

\begin{figure*}[ht!]
    \centering
    \includegraphics[width=1.0\textwidth]{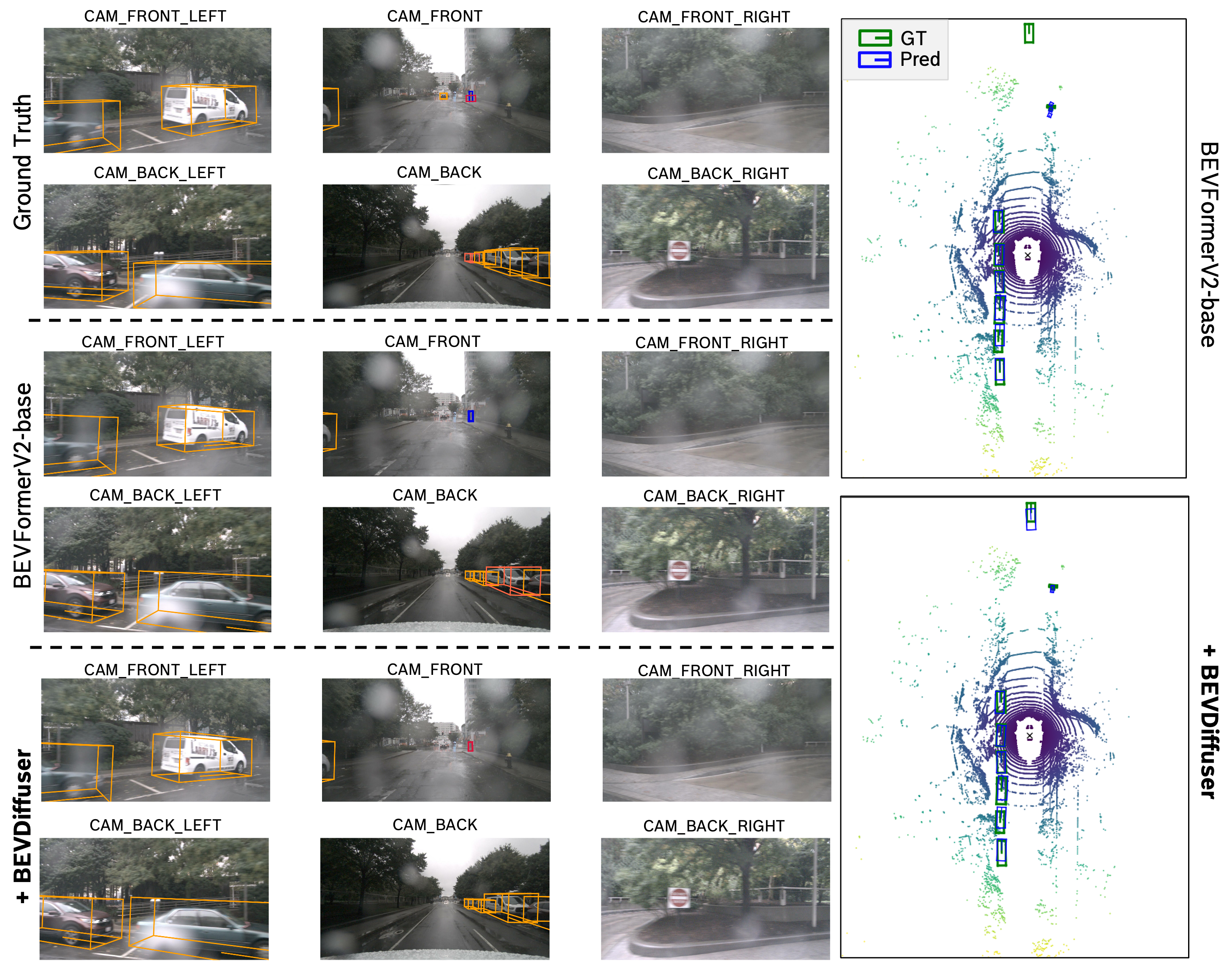}
    \caption{Visualization results of our BEVDiffuser enhanced BEVFormerV2 on nuScenes \texttt{val} dataset. In this representative example, despite the rain causing blurriness in the camera images, BEVDiffuser still enables BEVFormerV2 to reliably detect the object in front of the autonomous vehicle, as captured by the LiDAR top view.   }
    \label{fig:vis_v2}
\end{figure*}

\begin{figure*}[ht!]
    \centering
    \includegraphics[width=1.0\textwidth]{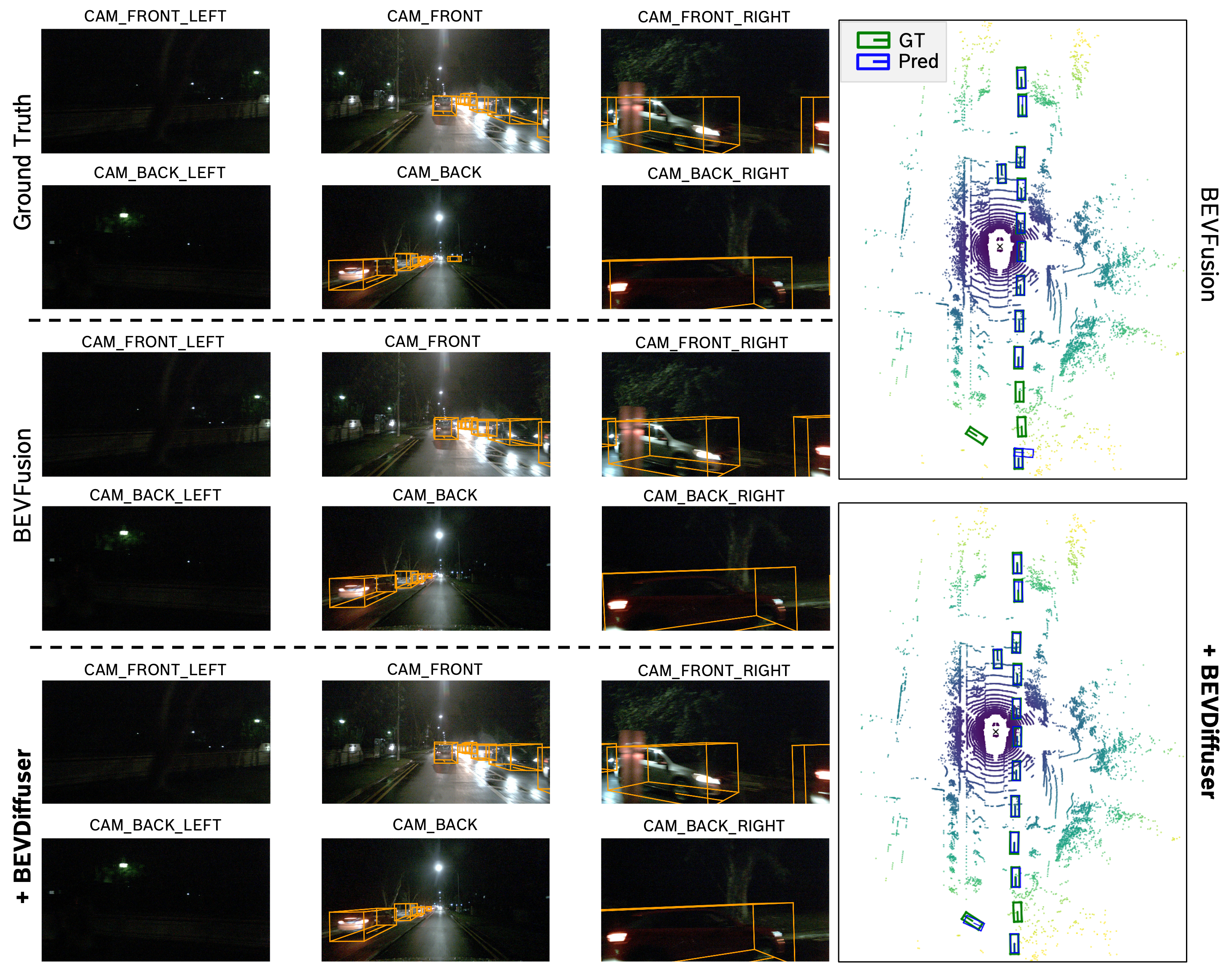}
    \caption{Visualization results of our BEVDiffuser enhanced BEVFusion on nuScenes \texttt{val} dataset. BEVFusion, which integrates both camera and LiDAR data, delivers robust performance in low-light conditions at night. BEVDiffuser further enhances BEVFusion by effectively reducing false negatives, as demonstrated in the LiDAR top view.}
    \label{fig:vis_fusion}
\end{figure*}

We visualize the 3D object detection results achieved by our BEVDiffuser enhanced BEVFormer-tiny, BEVFormer-base, BEVFormerV2 and BEVFusion in Fig.~\ref{fig:vis_tiny}, Fig.~\ref{fig:vis_base}, Fig.~\ref{fig:vis_v2} and Fig.~\ref{fig:vis_fusion}, respectively. We present the ground-truth and predicted 3D bounding boxes in both multi-camera images and the LiDAR top view to offer a comprehensive overview of the models' performance. As illustrated in the figures, BEVDiffuser consistently enhances the existing BEV models for object detection in complex environments and under challenging conditions by minimizing both false positives and false negatives, demonstrating its ability to improve the quality of the BEV representations.

\end{document}